\documentclass[runningheads]{llncs}

\usepackage{eccv}

\usepackage{eccvabbrv}
\usepackage{graphicx}
\usepackage{booktabs}
\usepackage{multirow}
\usepackage[accsupp]{axessibility}
\usepackage{hyphenat}

\usepackage[breaklinks,citecolor=eccvblue]{hyperref}
\usepackage[capitalize]{cleveref}
\usepackage{orcidlink}
\usepackage{listings}
\usepackage{mystyles}
\usepackage{marvosym}
\newcommand{\corrmark}{\textsuperscript{(\Letter)}}

\begin{document}

\title{Semantic Synthesis over Brute-Force Search: LLM-Guided Transformation Design \\ for Efficient Learning}
\titlerunning{Semantic Synthesis over Brute-Force Search: LLM-Guided Transformation}

\author{Usha Shrestha\corrmark\orcidlink{0009-0004-3224-4082}\and
Radu Timofte\orcidlink{0000-0002-1478-0402} \and Dmitry Ignatov\orcidlink{0000-0002-6339-1200}}
\authorrunning{U.~Shrestha et al.}
\institute{Computer Vision Lab, CAIDAS \& IFI, University of W\"urzburg, Germany\\ \email{ushaqshrestha@gmail.com, \{radu.timofte, dmytro.ignatov\}@uni-wuerzburg.de}}
\renewcommand{\thefootnote}{}
\footnotetext{\textsuperscript{\Letter} Corresponding author: \href{mailto:ushaqshrestha@gmail.com}{ushaqshrestha@gmail.com}}
\renewcommand{\thefootnote}{\arabic{footnote}}\setcounter{footnote}{0}
\maketitle

\begin{abstract}
Data transformation pipelines shape the training distribution in computer vision, but designing them is a computationally expensive problem. We investigate whether large language models can enable efficient discovery by synthesizing executable transformation programs outside predefined augmentation spaces. We compare nine strategies under a unified training protocol, including standard augmentation policies, brute-force search, direct code generation with two LLMs, and performance-guided generation using LLMs fine-tuned on accuracy-ordered transformation pairs. When the best candidates are retrained for 20 epochs, LLM-generated
transformation programs lead on every dataset, reaching 0.8877 accuracy on CIFAR-10, 0.5825 on CIFAR-100, and 0.9588 on SVHN, surpassing brute-force search over 6{,}000 candidates. The best transform on CIFAR-10 and SVHN comes from a fine-tuned generator using 270 candidates, while the best on CIFAR-100 comes from an off-the-shelf generator using only 10. This needs 22$\times$ fewer candidate evaluations in the generation step and 600$\times$ fewer for off-the-shelf generations, relative to brute-force search. Beyond transformation discovery, we analyze the reliability of the reduced-budget evaluation commonly used for augmentation search. Correlating one-epoch selection scores with 20-epoch outcomes reveals strong, weak, and absent predictive relationships depending on the dataset and architecture, with correlations ranging from 0.756 to near zero. A positive control indicates that range restriction alone is unlikely to explain the low correlations. Our results indicate that semantic synthesis with code LLMs provides an efficient alternative to search-based transformation discovery and highlight the importance of validating evaluation fidelity when optimizing data transformations.

\end{abstract}

\section{Introduction}
\label{sec:intro}

A neural network's accuracy depends on its architecture and on how its training data is preprocessed. Which augmentation pipeline to use for a given task has no settled answer, so practitioners choose among strategies that differ by orders of magnitude in cost: fixed policies such as RandAugment~\cite{randaugment} and TrivialAugment~\cite{trivialaugment} that require no search; meta-learned assistance that recommends preprocessing for a given algorithm~\cite{bilalli2018}; automated search such as AutoAugment~\cite{autoaugment}, whose policy space is large enough that reinforcement learning was needed to traverse it; and, recently, code generated by large language models~\cite{ding2024llmaug,duru2024adaptive}.

A transformation program is a curated data artifact. Viewed at the level of the training set rather than the model, a transformation program belongs with pruning, distillation, and synthetic data generation, and occupies a distinct position among them. Pruning removes existing samples; distillation learns a small synthetic replacement set; generative synthesis adds samples from a learned model. A transformation program does none of these: it edits the preprocessing distribution applied to every sample of an existing set, and only the code is produced and stored. The asymmetry is severe. The programs here are under $2$\,kB of source each, against $153.6$\,MB for CIFAR-10 as uint8 and $1.2$ to $6.1$\,MB for a distilled set at $10$ to $50$ images per class in float32 --- three orders of magnitude below distillation and five below the raw set, while acting on all $50{,}000$ samples. What such a program does commit to is a search procedure, and that procedure is what we examine here.

Most automated methods share a dependency that is rarely examined. Search requires scoring thousands of candidates; full training runs are too expensive to repeat thousands of times, so candidates are scored by a cheap proxy: train for a few epochs, or one epoch, and rank by validation accuracy. This proxy is standard practice in AutoML, but augmentation search inherits the practice without inheriting the validation that the pruning literature performs on its own proxies~\cite{coleman2020}.

We compare nine strategies under a single frozen training protocol on CIFAR-10, CIFAR-100~\cite{cifar}, and SVHN~\cite{svhn}, score each candidate after one epoch, then retrain the best candidates from each strategy for $20$ epochs, and compare the two orderings. Correlating each retrained candidate's one-epoch score against its own $20$-epoch accuracy gives $0.756$ on SVHN, $0.470$ on CIFAR-10 with MobileNetV2~\cite{mobilenetv2}, and values near zero on CIFAR-10 and CIFAR-100 with ResNet~\cite{resnet}. A positive control on CIFAR-10 checks the measurement directly: two independently trained one-epoch scores, restricted to the same narrow range as the candidates above, still correlate at $\rho = 0.816$. A strong relationship survives that range restriction, so the low correlations above are unlikely to be an artifact of it.

The language models prevail on all three datasets by a wide margin over brute-force search. The best transform found is $0.8877$ accuracy on CIFAR-10, $0.5825$ on CIFAR-100, and $0.9588$ on SVHN, compared to $0.8115$, $0.5027$, and $0.9394$ for brute force. That is $7.6$, $8.0$, and $1.9$ accuracy points, and brute force required $6{,}000$ trained models per dataset to reach its number, while the fine-tuned generators required $270$ on top of the corpus that seeds them and the off-the-shelf generators only $10$. Free-form code generation reaches configurations that a fixed operation dictionary cannot express, and it does so without searching over the dictionary.

This study builds on our previous single-dataset pipeline~\cite{shrestha2026brute}, which introduced the brute-force corpus construction and the performance-guided fine-tuning loop used here, and which was evaluated on CIFAR-10 with one base model on a one-epoch budget. We reuse that pipeline and extend it along each of these three axes: dataset, base model, and training length.

Our contributions are a performance-annotated corpus of $24{,}000$ evaluated PyTorch transformation programs across three datasets and two architectures, of which $18{,}000$ are new and the remaining $6{,}000$ are the CIFAR-10 corpus released with our previous work~\cite{shrestha2026brute}; $181$ full-budget retrainings with complete accuracy curves, which is what makes the proxy validation possible; and the comparison above, with an explicit account of how much of its margin the selection rule can explain.

\section{Related Work}
\label{sec:related}

\subsubsection{Automated augmentation search.}
AutoAugment~\cite{autoaugment} framed augmentation design as reinforcement-learning search over a discrete space of standard image operations and established the cost problem the field has worked on since. Fast AutoAugment~\cite{fastautoaugment} replaced policy training with density matching. RandAugment~\cite{randaugment} collapsed the space to two hyperparameters and matched AutoAugment's accuracy, evidence that the diversity of the transform vocabulary matters more than the sophistication of the search over it. TrivialAugment~\cite{trivialaugment} went further by removing search entirely, sampling a single transform uniformly at random per image, and targeted studies of individual operations continue to report task-specific gains that a fixed vocabulary does not anticipate~\cite{aboudeshish2025augmentest}. Others learn policies adversarially during training~\cite{advautoaugment}, through teacher guidance~\cite{teachaugment}, through differentiable relaxation~\cite{dada}, per-instance via reinforcement learning~\cite{minh2021instance}, or as nonstationary schedules over the course of training~\cite{pba}. All of these search within a fixed operational vocabulary, and all of them rank candidates with reduced fidelity. Reported evaluation budgets vary, and the fidelity used for ranking is often stated only in passing.

\subsubsection{Data curation and synthetic data.}
Curation methods differ in which object they modify. \emph{Pruning} discards samples, scoring them by forgetting events~\cite{toneva2019}, early-training gradient norms~\cite{paul2021}, submodular core set
objectives~\cite{mirzasoleiman2020}, or embedding\hyp-space margins that break power-law scaling~\cite{sorscher2022}; at web scale the same question becomes filter design~\cite{gadre2023datacomp,fang2024dfn}. \emph{Distillation} instead synthesizes a small replacement set, by gradient matching~\cite{zhao2021datasetcondensationgradientmatching}, trajectory matching~\cite{cazenavette2022mtt}, feature alignment~\cite{wang2022cafe}, kernel ridge regression~\cite{nguyen2021kip}, or generative priors~\cite{cazenavette2023glad}, following~\cite{wang2018dd} and surveyed in~\cite{yu2023review,sachdeva2023survey}. \emph{Generative synthetic data} adds samples from outside the original set entirely~\cite{azizi2023,sariyildiz2023,he2023,tian2023stablerep}. A
transformation program is a fourth object. The connection is methodological as well as conceptual: pruning has already
confronted the question this paper asks of augmentation search, since Coleman et al.~\cite{coleman2020} select data with a cheap proxy model and validate the proxy's ranking against full training explicitly. Augmentation search adopted reduced-fidelity ranking without a comparable validation, and the corpus released here is intended to supply the missing substrate.

\subsubsection{LLM-guided augmentation.}
LLMs have been proposed as augmentation engines~\cite{ding2024llmaug,mumuni2025autodata}. Closest to this work, Duru and Temizel~\cite{duru2024adaptive} couple LLM generation with downstream performance feedback, refining policies in context from a frozen model. In our previous work~\cite{shrestha2026brute}, we instead update the generator's weights on pairs ordered by measured accuracy, reporting that a fine-tuned OlympicCoder-7B matches brute-force peak accuracy on CIFAR-10 at a fraction of the evaluation count. Related work in the same framework studies few-shot prompting for architecture generation~\cite{vysyaraju2025fewshot}, whether an LLM produces novel rather
than memorized designs~\cite{khalid2026creativity}, and how the reference models
supplied to an LLM should be selected~\cite{ABrain.Analog}.

\subsubsection{LLMs as optimizers over code.}
A wider body of work uses LLMs as search agents guided by empirical scores.
OPRO~\cite{yang2023opro} conditions generation on sorted solution-score history,
FunSearch~\cite{funsearch} couples program synthesis with an evaluator,
EvoPrompting~\cite{evoprompting} applies evolutionary prompting to code-level
architecture search, Eureka~\cite{eureka} evolves reward code against
reinforcement-learning outcomes, and AlphaEvolve~\cite{alphaevolve} scales the
paradigm. These adapt in context or by evolutionary selection while the weights stay fixed; the method evaluated here distills feedback into weights by supervised pairs, with no reward model or ranking loss~\cite{bai2022rlhf}. Within the same framework, related work varies that supervised signal rather than the search loop, targeting code diffs instead of whole programs~\cite{ABrain.Delta} or ordering the training pairs by a similarity curriculum~\cite{ABrain.Similarity}; the signal used here is accuracy order alone. Kochnev et al.~\cite{kochnev2025optuna} instead benchmark an LLM against Optuna for hyperparameter tuning, the same head-to-head comparison this paper runs between discovery strategies rather than tuning algorithms.

\subsubsection{Instruction tuning for code.}
Supervised fine-tuning aligns models with task intent more effectively than scale alone~\cite{instructgpt,flan,parthasarathy2024}, code-corpus optimization improves functional correctness~\cite{codex}, and data quality outweighs volume~\cite{flan}. Guided generation methods such as AceCoder~\cite{acecoder} and LAIL~\cite{lail} use auxiliary signals to shape training examples; the signal here is downstream accuracy and nothing else. Prompt structure affects generation quality~\cite{kojima2022,sahoo2025prompt,schulhoff2025}.

\subsubsection{Reduced-fidelity evaluation.}
Evaluating candidates cheaply is standard in AutoML~\cite{successivehalving,hyperband,bohb,egele2024,ru2021}, and zero-cost proxies push the idea to its limit while acknowledging rank-correlation ceilings~\cite{zerocost,mellor2021}. That literature reports correlation against full training, and NAS-Bench-201~\cite{nasbench201} exists so such correlations can be measured at all. Closest to the question we pose, Adhikari et al.~\cite{ABrain.Converg} derive convergence conditions for iterative LLM-based search in which a closed-form proxy-reliability term determines whether the loop improves at all; the correlations reported in \cref{sec:proxy} measure the corresponding quantity empirically for transformation search. Augmentation search otherwise has no comparable substrate to our knowledge.

\section{Method}
\label{sec:method}

\subsection{Discovery Strategies}
\label{sec:strategies}

We use nine strategies in four families, all implemented under a single frozen protocol (\cref{sec:setup}).
\emph{Fixed policies} apply AutoAugment~\cite{autoaugment}, RandAugment~\cite{randaugment}, TrivialAugment~\cite{trivialaugment}, and a ``no augmentation'' control that applies only \texttt{Resize}, \texttt{ToTensor}, and \texttt{Normalize} operations, each at library defaults and contributing one candidate, a budget asymmetry we return to in \cref{sec:limitations}.

\emph{Brute-force search} samples compositions from a torchvision transform dictionary, $6{,}000$ candidates per dataset (\cref{sec:corpus}).

\emph{Off-the-shelf generation} prompts OlympicCoder-7B~\cite{olympiccoder} and Qwen2.5-Coder-7B~\cite{qwencoder} to generate transformation code, producing $10$ candidates per model for each dataset without any fine-tuning.

\begin{figure}[tb]
  \centering
  \includegraphics[width=0.75\linewidth]{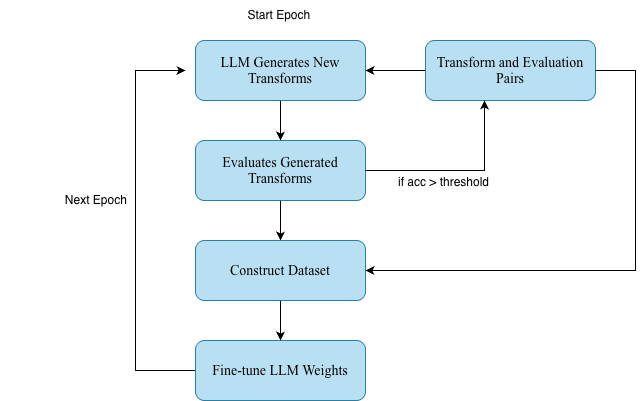}
  \caption{The fine-tuning loop. Each cycle samples
    reference transforms from the annotated pool, generates candidates (Listing~\ref{lst:gen}), evaluates them under the one-epoch protocol, admits those above a per-dataset threshold, and continues LoRA training on the rebuilt pair set. The loop runs $27$ cycles.}
  \label{fig:pipeline}
\end{figure}

\emph{Performance-fine-tuned generation} runs the same two models through an iterative loop that alternates between generation, evaluation, and LoRA training on accuracy-ordered pairs (see \cref{fig:pipeline}), generating $10$ candidates per cycle, per model, per dataset over $27$ cycles.

\subsection{Brute-Force Corpus}
\label{sec:corpus}

We follow our own corpus construction~\cite{shrestha2026brute} without modification so that the additional corpora are directly comparable to the CIFAR-10 corpus we released. Torchvision~\cite{torchvision} transforms are organized in a dictionary, with parameter randomization applied to each entry. Each candidate consists of $N \in \{1,2,3\}$ variable transforms, sampled as ordered $N$-permutations to account for the impact of composition order, followed by the fixed \texttt{Resize((64,64))}, \texttt{ToTensor}, and \texttt{Normalize} operations. There are $2{,}000$ candidates for each value of $N$ per dataset, resulting in $6{,}000$ candidates per corpus and $24{,}000$ across the four corpora in \Cref{tab:brute}. Each transform function is annotated with the accuracy it achieves under the one-epoch protocol of \cref{sec:setup}; no heuristic scores are used. Candidates that fail to parse or encounter runtime errors are assigned an accuracy of $0.0$ and remain in the corpus. Infrastructure failures are retried and not annotated.

\subsection{Performance-Guided Fine-Tuning}
\label{sec:finetune}

The fine-tuning loop follows the procedure we introduced in~\cite{shrestha2026brute}. From the annotated pool, we build conditional improvement pairs: for an anchor transform $A$, we sample transform $B$ with $\mathrm{acc}(B) > \mathrm{acc}(A)$ and $\mathrm{code}(B) \neq \mathrm{code}(A)$. The prompt (Listing~\ref{lst:ft}) presents $A$'s code and accuracy with an instruction to produce something better; the output is $B$'s code inside \texttt{<tr>} tags.

\begin{lstlisting}[language=JSON, basicstyle=\ttfamily\scriptsize, breaklines=true,
  caption={Prompt used for fine-tuning}, label={lst:ft}]
 "prompt": [
      "You are an expert image transformation optimizer.",
      "Baseline transform code (Accuracy: {accuracy}):",
      "<tr>{transform_code}</tr>",
      "Generate an improved Python transform function ('transform') that achieves a higher accuracy with ResNet-18, 1 epoch, batch 64, lr 0.01, dropout 0.2, and momentum 0.9 for '{dataset}' dataset and task: 'img-classification' ",
      "Your response MUST contain exactly one set of the XML tags <tr>...</tr>. DO NOT include any leading or trailing text, markdown fences (```), comments, or any other XML tags like <path> or <text>."
    ],
    "output": [
      "<tr>{addon_transform_code}</tr>"
    ]
\end{lstlisting}

\begin{lstlisting}[language=JSON, basicstyle=\ttfamily\scriptsize, breaklines=true,
  caption={Prompt used for generation}, label={lst:gen}]
"prompt": [
      "You are an expert image transformation generator.",
      "Your task is to generate new image transformation code.",
      "Use common patterns and ideas from the following two reference transforms:",
      "Reference 1 (Accuracy: {accuracy}):",
      "<tr>{transform_code}</tr>",
      "Reference 2 (Accuracy: {addon_accuracy}):",
      "<tr>{addon_transform_code}</tr>",
      "Provide a new, high-performance transform for '{dataset}' (task: 'img-classification') for training with ResNet-18, 1 epoch, batch 64, lr 0.01, dropout 0.2, and momentum 0.9",
      "Respond with:",
      "1. A <tr> XML tag containing the complete Python transform code (function name 'transform').",
      "The code must be wrapped strictly in <tr> and </tr> tags."
    ]
\end{lstlisting}

The seed set holds $3{,}541$ deduplicated transforms and their results for CIFAR-10, $3{,}518$ for CIFAR-100, and $3{,}519$ for SVHN, including constructed samples with \texttt{Resize} set to $256$. Each cycle samples $10$ anchor-reference prompts, generates $10$ candidates at temperature $0.8$, top-$k$ $70$, top-$p$ $0.9$, evaluates them, admits those above a per-dataset threshold into the pool, rebuilds the pair set, and continues LoRA training for three optimizer epochs. The loop runs $27$ cycles, so each fine-tuned model proposes $270$ candidates per dataset.

LoRA adapters use rank $32$, $\alpha = 32$, dropout $0.05$ on the attention and MLP projections of layers 0--23, trained with paged 8-bit AdamW under a cosine schedule at effective batch size $8$. Base models load in 4-bit NF4 so a 7B model and its optimizer fit on one 24\,GB card. Generated candidates are deduplicated against the pool by whitespace-normalized hash.

\section{Experimental Setup}
\label{sec:setup}

Fine-tuning of LLMs and training of computer vision models are performed using the AI $\text{Linux}$ docker image $\texttt{abrainone/ai-linux}$\footnote{AI Linux: \url{https://hub.docker.com/r/abrainone/AI-Linux}} on NVIDIA GeForce RTX 3090/4090 $24\text{GB}$ GPUs of the $\text{CVL Kubernetes}$ cluster at the University of Würzburg and a dedicated workstation. All strategies run inside the NNGPT\footnote{NNGPT: \url{https://github.com/ABrain-One/NN-GPT}} framework~\cite{nngpt}, and every transform is a PyTorch~\cite{pytorch} program executed against the same data loaders. Transforms are evaluated on image classification with the best-performing ResNet-18 variant stored in the LEMUR\footnote{LEMUR NN Dataset: \url{https://github.com/ABrain-One/NN-Dataset}} database~\cite{lemur,lemur2} for each dataset, trained from scratch at batch size $64$, learning rate $0.01$, momentum $0.9$, dropout $0.2$. Datasets are CIFAR-10, CIFAR-100~\cite{cifar}, and SVHN~\cite{svhn}.

Every candidate from every strategy is scored at one epoch at a flat learning rate of $0.01$. This is the selection budget, and the budget at which we reported results in~\cite{shrestha2026brute}.

We then take the top-$10$ candidates per search-based strategy per dataset by that score and retrain each from scratch for $20$ epochs with seed 42, cosine decay from $0.01$ to zero, logging test accuracy at every epoch. Fixed policies contribute one configuration and are retrained once. The $20$-epoch number is the outcome; the one-epoch number is the selection signal. After deduplication across strategies, this gives $43$ retrained candidates for CIFAR-10 with ResNet, $43$ for CIFAR-10 with MobileNetV2, $42$ for CIFAR-100, and $41$ for SVHN, which with the $4 \times 3$ fixed-policy runs makes the $181$ full-budget retrainings reported throughout.

Architecture sensitivity is probed on CIFAR-10 with MobileNetV2, both over the full $6{,}000$-candidate corpus at one epoch and over the $43$ retrained candidates at $20$ epochs (\cref{sec:arch}).

All correlations reported are Spearman rank correlations, denoted $\rho$, with two-sided $p$-values; confidence intervals are Fisher $z$ intervals using the variance-stabilized standard error $1/\sqrt{n-3}$ appropriate for Spearman's coefficient. The retrainings cost $80.7$ GPU-hours, with a median run of $14.4$ minutes on CIFAR-10, $18.7$ on CIFAR-100, and $35.6$ on SVHN.

\section{Results and Discussion}
\label{sec:results}

\subsection{The Brute-Force Corpus}

\Cref{tab:brute} reports the four corpora by composition length. Single-transform candidates exceed two- and three-transform compositions in mean accuracy, and the spread widens as compositions grow: under one epoch of training, stacking transforms mostly destabilizes optimization. Peak accuracy, which is the quantity search actually optimizes, is much flatter than the mean and almost invariant to $N$; on SVHN and on MobileNetV2 the best two-transform candidate edges past the best single one.

\begin{table}[tb]
\caption{Brute-force corpus by composition length, one-epoch evaluation. Candidates that failed to parse or execute are recorded at accuracy $0.0$ and are included in
$n$ and in the mean.}
\label{tab:brute}
\centering
\begin{tabular}{@{}lllcccc@{}}
\toprule
Dataset & Architecture & $N$ & $n$ & Mean & Best & 95\% CI (mean) \\
\midrule
\multirow{6}{*}{CIFAR-10}
 & \multirow{3}{*}{ResNet}      & 1 & 2000 & 0.5256 & \textbf{0.6124} & [0.5234, 0.5279] \\
 &                              & 2 & 2000  & 0.4832 & 0.6071 & [0.4801, 0.4863] \\
 &                              & 3 & 2000 & 0.4401 & 0.5983 & [0.4363, 0.4439] \\
\cmidrule(l){2-7}
 & \multirow{3}{*}{MobileNetV2} & 1 & 2000  & 0.3895 & 0.4604 & [0.3877, 0.3914] \\
 &                              & 2 & 2000 & 0.3569 & \textbf{0.4655} & [0.3545, 0.3593] \\
 &                              & 3 & 2000 & 0.3261 & 0.4517 & [0.3231, 0.3290] \\
\midrule
\multirow{3}{*}{CIFAR-100}
 & \multirow{3}{*}{ResNet}      & 1 & 2000 & 0.2004 & \textbf{0.2599} & [0.1988, 0.2019] \\
 &                              & 2 & 2000 & 0.1709 & 0.2526 & [0.1690, 0.1728] \\
 &                              & 3 & 2000  & 0.1466 & 0.2469 & [0.1445, 0.1488] \\
\midrule
\multirow{3}{*}{SVHN}
 & \multirow{3}{*}{ResNet}      & 1 & 2000  & 0.8375 & 0.8852 & [0.8360, 0.8390] \\
 &                              & 2 & 2000 & 0.8099 & \textbf{0.8872} & [0.8072, 0.8126] \\
 &                              & 3 & 2000 & 0.7584 & 0.8816 & [0.7540, 0.7628] \\
\bottomrule
\end{tabular}
\end{table}

\subsection{Selection at One Epoch}
\label{sec:oneepoch}

The Acc@1ep column of \Cref{tab:main} shows what the selection budget reports. Brute-force search wins CIFAR-10 ($0.6124$) and CIFAR-100 ($0.2599$) after $6{,}000$ evaluations each. On SVHN, the fine-tuned Qwen2.5-Coder-7B edges past it, $0.9063$ against $0.8872$, from $270$ candidates.

Fine-tuning helps here without exception, and the gain shows up on the one-epoch score used for selection. OlympicCoder gains $12.0$, $8.3$, and $6.8$ points over its base model on CIFAR-10, CIFAR-100, and SVHN, and Qwen gains $8.3$, $8.1$, and $17.5$. Fine-tuning reliably improves the LLM's performance on the objective it was trained on.

Fixed policies appear weak at this budget: on CIFAR-10 RandAugment ($0.3554$) and TrivialAugment ($0.4673$) fall below the no-augmentation control ($0.5832$). That is a property of the budget rather than of the methods, since strong stochastic augmentation slows early convergence and these policies target long schedules.

\begin{table}[tb]
\caption{Discovery strategies at both budgets, ResNet. Acc@1ep is the best one-epoch score over the strategy's search budget; Acc@20ep is its best candidate retrained for $20$ epochs. $\Delta$rank is the change in within-dataset rank between budgets, positive meaning the strategy improves its position at full budget.}
\label{tab:main}
\centering
\begin{tabular}{@{}llcccc@{}}
\toprule
Dataset & Strategy & Candidate & Acc@1ep & Acc@20ep & $\Delta$rank \\
\midrule
\multirow{9}{*}{CIFAR-10}
 & No augmentation     & 1    & 0.5832 & 0.7656 & $-3$ \\
 & AutoAugment         & 1    & 0.4083 & 0.7537 & $0$ \\
 & RandAugment         & 1    & 0.3554 & 0.7149 & $0$ \\
 & TrivialAugment      & 1    & 0.4673 & 0.8037 & $+1$ \\
 & Brute force         & 6000 & \textbf{0.6124} & 0.8115 & $-4$ \\
 & OlympicCoder-7B     & 10   & 0.4708 & 0.8728 & $+4$ \\
 & \quad + fine-tuning & 270  & 0.5903 & 0.8626 & $0$ \\
 & Qwen2.5-Coder-7B    & 10   & 0.5141 & 0.8140 & $+1$ \\
 & \quad + fine-tuning & 270  & 0.5970 & \textbf{0.8877} & $+1$ \\
\midrule
\multirow{9}{*}{CIFAR-100}
 & No augmentation     & 1    & 0.2350 & 0.4342 & $-3$ \\
 & AutoAugment         & 1    & 0.1373 & 0.4176 & $0$ \\
 & RandAugment         & 1    & 0.1193 & 0.3973 & $0$ \\
 & TrivialAugment      & 1    & 0.1522 & 0.4810 & $+1$ \\
 & Brute force         & 6000 & \textbf{0.2599} & 0.5027 & $-2$ \\
 & OlympicCoder-7B     & 10   & 0.1625 & 0.5444 & $+4$ \\
 & \quad + fine-tuning & 270  & 0.2453 & 0.4873 & $-2$ \\
 & Qwen2.5-Coder-7B    & 10   & 0.1649 & \textbf{0.5825} & $+4$ \\
 & \quad + fine-tuning & 270  & 0.2455 & 0.4913 & $-2$ \\
\midrule
\multirow{9}{*}{SVHN}
 & No augmentation     & 1    & 0.8580 & 0.9345 & $-4$ \\
 & AutoAugment         & 1    & 0.8077 & 0.9480 & $+4$ \\
 & RandAugment         & 1    & 0.6711 & 0.8826 & $0$ \\
 & TrivialAugment      & 1    & 0.8308 & 0.9474 & $+1$ \\
 & Brute force         & 6000 & 0.8872 & 0.9394 & $-3$ \\
 & OlympicCoder-7B     & 10   & 0.8268 & 0.9415 & $+1$ \\
 & \quad + fine-tuning & 270  & 0.8943 & 0.9541 & $0$ \\
 & Qwen2.5-Coder-7B    & 10   & 0.7312 & 0.9357 & $+1$ \\
 & \quad + fine-tuning & 270  & \textbf{0.9063} & \textbf{0.9588} & $0$ \\
\bottomrule
\end{tabular}
\end{table}

\subsection{Full-Budget Results}
\label{sec:20epoch}

At $20$ epochs a language model produces the best transform on every dataset. Qwen fine-tuned takes CIFAR-10 at $0.8877$ and SVHN at $0.9588$; Qwen off the shelf takes CIFAR-100 at $0.5825$.

Against brute-force search the margins are $7.6$, $8.0$, and $1.9$ accuracy points. Against the strongest non-LLM arm, the picture is not uniform. On CIFAR-10 and CIFAR-100 brute force is that arm ($0.8115$, $0.5027$), so both margins stand. On SVHN it is not: AutoAugment reaches $0.9480$, above brute force's $0.9394$, reducing the LLM margin over the best non-LLM baseline to $1.1$ points, which at a single seed is not separable from run-to-run variance. SVHN is where our headline claim is weakest.

The cost difference is larger than the accuracy difference. Brute force trained $6{,}000$ models per dataset; the fine-tuned generators $270$, the off-the-shelf generators $10$. The best LLM arm beats brute force on all three datasets at $22$ and $600$ times fewer evaluations, and the fixed policies, at one run each, land below every LLM arm everywhere except SVHN, where AutoAugment ($0.9480$) and TrivialAugment ($0.9474$) both sit between the two off-the-shelf models and the two fine-tuned ones.

Evaluation counts understate the asymmetry and also flatter the fine-tuned arm. Taking per-epoch cost as the median $20$-epoch run divided by $20$ ($0.72$, $0.94$, and $1.78$ minutes), the three ResNet brute-force searches account for roughly $72$, $94$, and $178$ GPU-hours of candidate evaluation, the $270$-candidate arms for $3.2$ to $8.0$ hours, and the $10$-candidate arms for $7$ to $18$ GPU-\emph{minutes}. The fine-tuned arm is not cheap in isolation: its pair set comes from the pool built out of the brute-force search, so those hours sit on top of the brute-force search cost rather than replacing it, and $27$ LoRA cycles plus $270$ generations from the 4-bit 7B model must be added. The off-the-shelf arm carries no such debt, and the best CIFAR-100 transform came from it: roughly $9$ GPU-minutes of candidate evaluation, no corpus and no fine-tuning, beating a $94$-GPU-hour brute-force search by $8.0$ points.

Fine-tuning holds on two of the three datasets. On SVHN both fine-tuned LLMs improve over their base versions, $0.9415$ to $0.9541$ for OlympicCoder and $0.9357$ to $0.9588$ for Qwen. On CIFAR-10 fine-tuned Qwen produces the best transform in the study, while fine-tuned OlympicCoder falls a point below its own base model. On CIFAR-100 neither fine-tuned model beats its base version. The pattern tracks proxy fidelity dataset by dataset (\cref{sec:base}).

Rank changes between the two budgets are mostly small. RandAugment is last at both budgets on CIFAR-10 and SVHN, and fine-tuned Qwen is first at both on SVHN. Brute force is the one strategy that moves consistently in the wrong direction, dropping four, two, and three places.

\subsection{How Well Does the Selection Score Predict the Outcome?}
\label{sec:proxy}

Every method in \Cref{tab:main} chooses its candidates using a one-epoch score, so it is worth asking what that score actually captures. \Cref{tab:rho} correlates each retrained candidate's own selection score against its own $20$-epoch accuracy.

The answer depends on the setting. SVHN gives $0.756$, interval $[0.58, 0.86]$: a usable signal. CIFAR-10 with MobileNetV2 gives $0.470$, $p = 0.002$, though its interval $[0.20, 0.68]$ spans marginal to strong. CIFAR-10 and CIFAR-100 with ResNet give $-0.024$ and $+0.024$, with intervals centered on zero that exclude any relationship stronger than moderate. At $n \approx 42$ these intervals are wide, and we read them as ruling out a strong proxy rather than as establishing exactly zero.

Because the retrained candidates were themselves selected by that score, a natural objection is that their scores are too similar for any correlation to appear. Panel B addresses this for CIFAR-10: two independently trained one-epoch scores, ResNet and MobileNetV2 on the same $6{,}000$ transforms, correlate at $0.853$ over the whole corpus and still at $0.816$ when restricted to the exact score span of Panel A, so a strong relationship is recoverable inside that span. Two caveats bound how far the control reaches. It relates two measurements of the same quantity type, one epoch to one epoch, which is easier than one epoch to $20$; and the restriction retains $5{,}548$ of $6{,}000$ candidates, matching Panel A's range but not its distribution, since the $43$ retrained candidates concentrate near the top of that range.

The signal is recoverable by training a little longer. Correlating accuracy at epoch $k$ of the $20$-epoch run against its endpoint gives $-0.243$, $-0.046$, and $0.738$ at $k = 1$ across the three ResNet settings, and $0.918$, $0.835$, and $0.882$ at $k = 10$. These epoch-$1$ figures are not the Panel A figures, which come from the independent flat-learning-rate selection run rather than the first epoch of a cosine schedule. More importantly, correlating epoch $k$ against epoch $20$ \emph{of the same run} shares initialization, data order, and schedule, so it measures trajectory autocorrelation and is optimistic about what an independent $k$-epoch proxy would achieve. Ten epochs is a promising fidelity, not a validated one.

We draw a narrow conclusion. One epoch carries real information in some dataset and architecture pairings and very little in others, and nothing in the score itself indicates which case one is in. That is a reason to report the fidelity used for ranking, and ideally its correlation against full training, alongside any augmentation-search result.

\begin{table}[tb]
\caption{How well the one-epoch selection score predicts the $20$-epoch outcome. Panel A correlates each retrained candidate's own selection score against its own final accuracy, over the score span given. Panel B is a positive control: two independently trained one-epoch scores on CIFAR-10, over the whole corpus and then restricted to Panel A's span. Coefficients are Spearman $\rho$ with $95\%$ Fisher $z$ intervals.}
\label{tab:rho}
\centering
\small
\setlength{\tabcolsep}{4pt}
\begin{tabular}{@{}llccccc@{}}
\toprule
Dataset & Architecture & $n$ & Score span & $\rho$ & $95\%$ CI & $p$ \\
\midrule
\multicolumn{7}{@{}l}{\emph{Panel A: one-epoch selection score $\rightarrow$ 20-epoch accuracy}} \\
CIFAR-10  & ResNet      & 43 & 0.350--0.612 & $-0.024$ & $[-0.32, +0.28]$ & 0.88 \\
CIFAR-10  & MobileNetV2 & 43 & 0.200--0.527 & $+0.470$ & $[+0.20, +0.68]$ & 0.002 \\
CIFAR-100 & ResNet      & 42 & 0.035--0.260 & $+0.024$ & $[-0.28, +0.33]$ & 0.88 \\
SVHN      & ResNet      & 41 & 0.524--0.906 & $+0.756$ & $[+0.58, +0.86]$ & $<$0.001 \\
\midrule
\multicolumn{7}{@{}l}{\emph{Panel B: control, ResNet one-epoch $\rightarrow$ MobileNetV2 one-epoch}} \\
CIFAR-10 & both & 6000 & full corpus  & $+0.853$ & $[+0.846, +0.860]$ & $<$0.001 \\
CIFAR-10 & both & 5548 & 0.350--0.612 & $+0.816$ & $[+0.807, +0.825]$ & $<$0.001 \\
\bottomrule
\end{tabular}
\end{table}

\subsection{Where the Evaluation Budget Goes}
\label{sec:budget}

Aggregating each dataset's five candidate-generating arms by mean one-epoch
selection score against mean $20$-epoch outcome reproduces the split above at the
arm level. Where the per-candidate correlation is strong the two orderings agree:
on SVHN ($\rho_{\text{cand}} = 0.756$) arm-level score and outcome nearly
coincide ($\rho_{\text{arm}} = +0.90$), with fine-tuned Qwen leading both and
brute force mid-table on both. CIFAR-100 with ResNet and CIFAR-10 with
MobileNetV2 sit between the extremes ($+0.20$ and $+0.30$). Where the
per-candidate correlation is near zero the orderings invert: on CIFAR-10 with
ResNet ($\rho_{\text{cand}} = -0.024$), $\rho_{\text{arm}} = -0.60$, and
brute-force search achieves the highest mean selection score together with the
lowest mean retrained outcome. Each arm-level coefficient is computed over the
five candidate-generating arms of its setting and is reported as a descriptive
summary rather than as a test.

This disparity exemplifies the exploitation failure documented in offline
model-based optimization~\cite{designbench}. When an optimizer
selects the absolute peak from thousands of noisy, reduced-fidelity
evaluations, it systematically concentrates on candidates that the
surrogate most overrates.

LLM generation structurally avoids this trap by drawing only a few candidates. Because the models generate a restricted pool between $10$ and $270$ candidates, they are never exposed to an argmax over thousands of noisy proxies. The language models succeed not merely by proposing better operations, but by bypassing the proxy overfitting inherent to large-scale brute-force search under severe budget constraints.

Separating the effect of the candidate source from the effect of the selection rule requires a random-subset control: drawing candidates uniformly from the brute-force corpus at the size of a generator budget, retraining them for $20$ epochs without any proxy-based selection, and comparing the resulting outcome distribution against the generated arms. We did not run that control, and \cref{sec:limitations} records the confound it would resolve.

\subsection{Base Model}
\label{sec:base}

Earlier work used a single generator, so whether its results were a property of the method or of OlympicCoder-7B was untestable. Adding Qwen2.5-Coder-7B addresses this. Both models improve their one-epoch output under fine-tuning on all three datasets, and both produce $20$-epoch transforms that beat brute force. On mean $20$-epoch accuracy, SVHN gains on both ($0.9296$ to $0.9423$, $0.9126$ to $0.9556$), CIFAR-100 gains on neither, and CIFAR-10 splits. Two base models moving together, dataset by dataset, is the main reason we read that pattern as a property of the training signal rather than of one model.

The direction of that pattern follows the proxy. Fine-tuning is trained on pairs ordered by the one-epoch score, so it can improve full-budget outcomes only where that ordering is faithful: it gains on SVHN, where $\rho = 0.756$, and not on CIFAR-100, where $\rho \approx 0$. This is the dependence that convergence analyses of iterative LLM-based search make explicit~\cite{ABrain.Converg}. Fine-tuning did not fail there; it optimized the objective it was given.

\subsection{Architecture Transfer}
\label{sec:arch}

Transforms discovered on one network transfer to another. Over the full $6{,}000$-candidate CIFAR-10 corpus at one epoch, ResNet and MobileNetV2 accuracies correlate at $0.853$, CI $[0.846, 0.860]$. Over the $43$ retrained candidates at $20$ epochs they correlate at $0.855$, CI $[0.75, 0.92]$, with MobileNetV2 lower by $3.6$ points on average. Two very different measurements, at different budgets and sample sizes, agree.

The arm ordering survives too. On MobileNetV2 the fine-tuned OlympicCoder arm leads on mean selection score and the best transform again comes from fine-tuned Qwen, at $0.8759$, so the winning arm is the one that also wins under ResNet. Brute force ranks second on mean selection score and last on outcome, repeating under MobileNetV2 the outcome position it holds under ResNet. The architecture used during search can therefore be chosen for speed; it is the training length, not the network, that has to be paid for.

\section{Limitations}
\label{sec:limitations}

Every accuracy here is one run at seed $42$. The smaller gaps in \Cref{tab:main} are not separable from seed variance, and the correlations inherit that noise. The claims we build on are the large ones: a $7.6$-point margin on CIFAR-10, an $8.0$-point margin on CIFAR-100, and correlations measured over $41$ to $6{,}000$ points.

The fixed policies are evaluated at library defaults and contribute one candidate each against $10$ to $270$ for the search arms, and RandAugment's two hyperparameters were not tuned; a matched-budget grid would be the fair comparison. The brute-force versus generation comparison confounds candidate distribution with selection rule (\cref{sec:budget}).

On the proxy analysis, the positive control relates one-epoch to one-epoch measurements and matches Panel A's range but not its top-heavy distribution, and the epoch-$k$ analysis is measured within single runs; a stratified retraining spanning the full score range rather than the selected top of it would settle both. Finally, $20$ epochs is far shorter than the schedules under which augmentation methods report published gains, so whether a $20$-epoch ranking predicts a $200$-epoch ranking is untested, and evaluation is restricted to convolutional architectures at fixed low resolution.

\section{Conclusion}

Two results are worth separating, and they carry different weight.

The stronger one concerns evaluation. The one-epoch score used to select candidates predicts the $20$-epoch outcome at $\rho = 0.756$ on SVHN and near zero on CIFAR-10 and CIFAR-100 with ResNet. A positive control indicates that range restriction alone is unlikely to explain this gap. Fine-tuning inherits it directly, since it trains on pairs ordered by that same score: results improve on SVHN, where the ordering holds, and stall on CIFAR-100, where it does not. What fails there is the objective supplied to the optimizer, not the optimization itself. Because augmentation search has no equivalent of NAS-Bench-201, these correlations have, to our knowledge, not previously been measured at this scale; the $24{,}000$-program corpus released here is meant to make them measurable for methods other than ours.

The weaker one concerns discovery. On all three datasets the best transform at $20$ epochs came from a language model, beating $6{,}000$ brute-force search candidates while training $22$ to $600$ times fewer models in the generation step. Free-form transform code generation reaches compositions and parameter settings that a fixed operation dictionary cannot express, and generating a handful of them is cheaper than searching the dictionary. We hold that at arm's length for the reason given in \cref{sec:budget}: brute-force search and generation differ in selection rule as well as in generator, and on the two datasets with the largest margins the selection rule alone predicts the direction of the outcome. What we are confident of concerns budget allocation rather than generators: taking the best of a few generated candidates is preferable to taking the best of thousands, because an argmax over thousands of noisy scores concentrates on whichever candidate the score most overrates, whereas an argmax over 10 is less exposed to that; and search may be conducted on the cheaper network, since ranking transfer reaches $0.85$.

Two inexpensive directions follow. The random-subset control described in \cref{sec:budget} would separate generator from selection rule. And since a transformation program costs kilobytes while acting on the whole training set, its value should be largest where samples are scarcest: evaluating the discovered transforms on $10\%$ and $25\%$ subsets would place transformation synthesis alongside pruning and distillation.

\vspace{0.4cm}
\noindent\textbf{Acknowledgments.}
This work was partially supported by the Alexander von Humboldt Foundation.

{\small
\bibliographystyle{splncs04}
\bibliography{eccv_paper}

@String(CVPR  = {IEEE Conf. Comput. Vis. Pattern Recog.})

@String(ECCV  = {Eur. Conf. Comput. Vis.})

@String(ICLR  = {Int. Conf. Learn. Represent.})

@String(CVPRW = {IEEE Conf. Comput. Vis. Pattern Recog. Worksh.})

@String(CVPR  = {CVPR})

@String(ECCV  = {ECCV})

@String(ICLR  = {ICLR})

@String(CVPRW = {CVPRW})

@inproceedings{toneva2019,
  title={An empirical study of example forgetting during deep neural network learning},
  author={Toneva, Mariya and Sordoni, Alessandro and Des Combes, Remi Tachet and Trischler, Adam and Bengio, Yoshua and Gordon, Geoffrey J},
  booktitle={International conference on learning representations},
  year={2018}
}

@article{paul2021,
  title={Deep learning on a data diet: Finding important examples early in training},
  author={Paul, Mansheej and Ganguli, Surya and Dziugaite, Gintare Karolina},
  journal={Advances in neural information processing systems},
  volume={34},
  pages={20596--20607},
  year={2021}
}

@article{sorscher2022,
  title={Beyond neural scaling laws: beating power law scaling via data pruning},
  author={Sorscher, Ben and Geirhos, Robert and Shekhar, Shashank and Ganguli, Surya and Morcos, Ari},
  journal={Advances in Neural Information Processing Systems},
  volume={35},
  pages={19523--19536},
  year={2022}
}

@inproceedings{mirzasoleiman2020,
  title={Coresets for data-efficient training of machine learning models},
  author={Mirzasoleiman, Baharan and Bilmes, Jeff and Leskovec, Jure},
  booktitle={International Conference on Machine Learning},
  pages={6950--6960},
  year={2020},
  organization={PMLR}
}

@article{coleman2020,
  title={Selection via proxy: Efficient data selection for deep learning},
  author={Coleman, Cody and Yeh, Christopher and Mussmann, Stephen and Mirzasoleiman, Baharan and Bailis, Peter and Liang, Percy and Leskovec, Jure and Zaharia, Matei},
  journal={arXiv preprint arXiv:1906.11829},
  year={2019}
}

@article{gadre2023datacomp,
  title={Datacomp: In search of the next generation of multimodal datasets},
  author={Gadre, Samir Yitzhak and Ilharco, Gabriel and Fang, Alex and Hayase, Jonathan and Smyrnis, Georgios and Nguyen, Thao and Marten, Ryan and Wortsman, Mitchell and Ghosh, Dhruba and Zhang, Jieyu and others},
  journal={Advances in Neural Information Processing Systems},
  volume={36},
  pages={27092--27112},
  year={2023}
}

@inproceedings{fang2024dfn,
  title={Data filtering networks},
  author={Fang, Alex and Madappally Jose, Albin and Jain, Amit and Schmidt, Ludwig and Toshev, Alexander and Shankar, Vaishaal},
  booktitle={International Conference on Learning Representations},
  volume={2024},
  pages={36221--36237},
  year={2024}
}

@article{wang2018dd,
  title={Dataset distillation},
  author={Wang, Tongzhou and Zhu, Jun-Yan and Torralba, Antonio and Efros, Alexei A},
  journal={arXiv preprint arXiv:1811.10959},
  year={2018}
}

@inproceedings{cazenavette2022mtt,
  title={Dataset distillation by matching training trajectories},
  author={Cazenavette, George and Wang, Tongzhou and Torralba, Antonio and Efros, Alexei A and Zhu, Jun-Yan},
  booktitle={Proceedings of the IEEE/CVF conference on computer vision and pattern recognition},
  pages={4750--4759},
  year={2022}
}

@inproceedings{cazenavette2023glad,
  title={Generalizing dataset distillation via deep generative prior},
  author={Cazenavette, George and Wang, Tongzhou and Torralba, Antonio and Efros, Alexei A and Zhu, Jun-Yan},
  booktitle={Proceedings of the IEEE/CVF conference on computer vision and pattern recognition},
  pages={3739--3748},
  year={2023}
}

@article{nguyen2021kip,
  title={Dataset meta-learning from kernel ridge-regression},
  author={Nguyen, Timothy and Chen, Zhourong and Lee, Jaehoon},
  journal={arXiv preprint arXiv:2011.00050},
  year={2020}
}

@inproceedings{wang2022cafe,
  title={Cafe: Learning to condense dataset by aligning features},
  author={Wang, Kai and Zhao, Bo and Peng, Xiangyu and Zhu, Zheng and Yang, Shuo and Wang, Shuo and Huang, Guan and Bilen, Hakan and Wang, Xinchao and You, Yang},
  booktitle={Proceedings of the IEEE/CVF Conference on Computer Vision and Pattern Recognition},
  pages={12196--12205},
  year={2022}
}

@article{yu2023review,
  title={Dataset distillation: A comprehensive review},
  author={Yu, Ruonan and Liu, Songhua and Wang, Xinchao},
  journal={IEEE transactions on pattern analysis and machine intelligence},
  volume={46},
  number={1},
  pages={150--170},
  year={2023},
  publisher={IEEE}
}

@article{sachdeva2023survey,
  title={Data distillation: A survey},
  author={Sachdeva, Noveen and McAuley, Julian},
  journal={arXiv preprint arXiv:2301.04272},
  year={2023}
}

@article{azizi2023,
  title={Synthetic data from diffusion models improves imagenet classification},
  author={Azizi, Shekoofeh and Kornblith, Simon and Saharia, Chitwan and Norouzi, Mohammad and Fleet, David J},
  journal={arXiv preprint arXiv:2304.08466},
  year={2023}
}

@inproceedings{sariyildiz2023,
  title={Fake it till you make it: Learning transferable representations from synthetic imagenet clones},
  author={Sariyildiz, Mert Bulent and Alahari, Karteek and Larlus, Diane and Kalantidis, Yannis},
  booktitle={CVPR 2023-IEEE/CVF conference on computer vision and pattern recognition},
  pages={8011--8021},
  year={2023},
  organization={IEEE}
}

@article{he2023,
  title={Is synthetic data from generative models ready for image recognition?},
  author={He, Ruifei and Sun, Shuyang and Yu, Xin and Xue, Chuhui and Zhang, Wenqing and Torr, Philip and Bai, Song and Qi, Xiaojuan},
  journal={arXiv preprint arXiv:2210.07574},
  year={2022}
}

@article{tian2023stablerep,
  title={Stablerep: Synthetic images from text-to-image models make strong visual representation learners},
  author={Tian, Yonglong and Fan, Lijie and Isola, Phillip and Chang, Huiwen and Krishnan, Dilip},
  journal={Advances in Neural Information Processing Systems},
  volume={36},
  pages={48382--48402},
  year={2023}
}

@inproceedings{autoaugment,
  title={Autoaugment: Learning augmentation strategies from data},
  author={Cubuk, Ekin D and Zoph, Barret and Mane, Dandelion and Vasudevan, Vijay and Le, Quoc V},
  booktitle={Proceedings of the IEEE/CVF conference on computer vision and pattern recognition},
  pages={113--123},
  year={2019}
}

@inproceedings{randaugment,
  title={Randaugment: Practical automated data augmentation with a reduced search space},
  author={Cubuk, Ekin D and Zoph, Barret and Shlens, Jonathon and Le, Quoc V},
  booktitle={Proceedings of the IEEE/CVF conference on computer vision and pattern recognition workshops},
  pages={702--703},
  year={2020}
}

@inproceedings{trivialaugment,
  title={Trivialaugment: Tuning-free yet state-of-the-art data augmentation},
  author={M{\"u}ller, Samuel G and Hutter, Frank},
  booktitle={Proceedings of the IEEE/CVF international conference on computer vision},
  pages={774--782},
  year={2021}
}

@article{fastautoaugment,
  title={Fast autoaugment},
  author={Lim, Sungbin and Kim, Ildoo and Kim, Taesup and Kim, Chiheon and Kim, Sungwoong},
  journal={Advances in neural information processing systems},
  volume={32},
  year={2019}
}

@inproceedings{pba,
  title={Population based augmentation: Efficient learning of augmentation policy schedules},
  author={Ho, Daniel and Liang, Eric and Chen, Xi and Stoica, Ion and Abbeel, Pieter},
  booktitle={International conference on machine learning},
  pages={2731--2741},
  year={2019},
  organization={PMLR}
}

@inproceedings{dada,
  title={Differentiable automatic data augmentation},
  author={Li, Yonggang and Hu, Guosheng and Wang, Yongtao and Hospedales, Timothy and Robertson, Neil M and Yang, Yongxin},
  booktitle={European conference on computer vision},
  pages={580--595},
  year={2020},
  organization={Springer}
}

@article{advautoaugment,
  title={Adversarial autoaugment},
  author={Zhang, Xinyu and Wang, Qiang and Zhang, Jian and Zhong, Zhao},
  journal={arXiv preprint arXiv:1912.11188},
  year={2019}
}

@inproceedings{teachaugment,
  title={Teachaugment: Data augmentation optimization using teacher knowledge},
  author={Suzuki, Teppei},
  booktitle={Proceedings of the IEEE/CVF conference on computer vision and pattern recognition},
  pages={10904--10914},
  year={2022}
}

@article{minh2021instance,
  title={Automated image data preprocessing with deep reinforcement learning},
  author={Minh, Tran Ngoc and Sinn, Mathieu and Lam, Hoang Thanh and Wistuba, Martin},
  journal={arXiv preprint arXiv:1806.05886},
  year={2018}
}

@inproceedings{duru2024adaptive,
  title={Adaptive augmentation policy optimization with {LLM} feedback},
  author={Duru, Ant and Temizel, Alptekin},
  booktitle={Eighteenth International Conference on Machine Vision (ICMV 2025)},
  volume={14114},
  pages={228--235},
  year={2026},
  organization={SPIE}
}

@inproceedings{ding2024llmaug,
    title = "Data Augmentation using {LLM}s: Data Perspectives, Learning Paradigms and Challenges",
    author = "Ding, Bosheng  and
      Qin, Chengwei  and
      Zhao, Ruochen  and
      Luo, Tianze  and
      Li, Xinze  and
      Chen, Guizhen  and
      Xia, Wenhan  and
      Hu, Junjie  and
      Luu, Anh Tuan  and
      Joty, Shafiq",
    editor = "Ku, Lun-Wei  and
      Martins, Andre  and
      Srikumar, Vivek",
    booktitle = "Findings of the Association for Computational Linguistics: ACL 2024",
    month = aug,
    year = "2024",
    address = "Bangkok, Thailand",
    publisher = "Association for Computational Linguistics",
    doi = "10.18653/v1/2024.findings-acl.97",
    pages = "1679--1705"
}

@inproceedings{yang2023opro,
  title={Large language models as optimizers},
  author={Yang, Chengrun and Wang, Xuezhi and Lu, Yifeng and Liu, Hanxiao and Le, Quoc V and Zhou, Denny and Chen, Xinyun},
  booktitle={International Conference on Learning Representations},
  volume={2024},
  pages={12028--12068},
  year={2024}
}

@article{funsearch,
  title={Mathematical discoveries from program search with large language models},
  author={Romera-Paredes, Bernardino and Barekatain, Mohammadamin and Novikov, Alexander and Balog, Matej and Kumar, M Pawan and Dupont, Emilien and Ruiz, Francisco JR and Ellenberg, Jordan S and Wang, Pengming and Fawzi, Omar and others},
  journal={Nature},
  volume={625},
  number={7995},
  pages={468--475},
  year={2024},
  publisher={Nature Publishing Group UK London}
}

@article{evoprompting,
  title={Evoprompting: Language models for code-level neural architecture search},
  author={Chen, Angelica and Dohan, David and So, David},
  journal={Advances in neural information processing systems},
  volume={36},
  pages={7787--7817},
  year={2023}
}

@article{eureka,
  title={Eureka: Human-level reward design via coding large language models},
  author={Ma, Yecheng Jason and Liang, William and Wang, Guanzhi and Huang, De-An and Bastani, Osbert and Jayaraman, Dinesh and Zhu, Yuke and Fan, Linxi and Anandkumar, Anima},
  journal={arXiv preprint arXiv:2310.12931},
  year={2023}
}

@article{alphaevolve,
  title={Alphaevolve: A coding agent for scientific and algorithmic discovery},
  author={Novikov, Alexander and V{\~u}, Ng{\^a}n and Eisenberger, Marvin and Dupont, Emilien and Huang, Po-Sen and Wagner, Adam Zsolt and Shirobokov, Sergey and Kozlovskii, Borislav and Ruiz, Francisco JR and Mehrabian, Abbas and others},
  journal={arXiv preprint arXiv:2506.13131},
  year={2025}
}

@article{shrestha2026brute,
  title={From Brute Force to Semantic Insight: Performance-Guided Data Transformation Design with {LLM}s},
  author={Shrestha, Usha and Ignatov, Dmitry and Timofte, Radu},
  journal={arXiv preprint arXiv:2601.03808},
  year={2026}
}

@inproceedings{nngpt,
  title={{NNGPT}: Rethinking {AutoML} with Large Language Models},
author={Kochnev, Roman and Khalid, Waleed and Uzun, Tolgay Atinc and Zhang, Xi and Dhameliya, Yashkumar Sanjaybhai and Qin, Furui and Vysyaraju, Chandini and Duvvuri, Raghuvir and Goyal, Avi and Ignatov, Dmitry and others},
  booktitle={Proceedings of the IEEE/CVF Conference on Computer Vision and Pattern Recognition},
  pages={3262--3272},
  year={2026}
}

@article{lemur,
  title={{LEMUR} Neural Network Dataset: Towards Seamless {AutoML}},
	author={Goodarzi, Arash Torabi and Kochnev, Roman and Khalid, Waleed and Qin, Furui and Uzun, Tolgay Atinc and Dhameliya, Yashkumar Sanjaybhai and Kathiriya, Yash Kanubhai and Bentyn, Zofia Antonina and Ignatov, Dmitry and Timofte, Radu},	
	primaryClass = {cs.LG},	
	journal = {arXiv preprint},
  	url = {https://arxiv.org/abs/2504.10552},
	year={2025}
}

@inproceedings{kochnev2025optuna,
  title={Optuna vs code LlaMa: Are LLMs a new paradigm for hyperparameter tuning?},
  author={Kochnev, Roman and Goodarzi, Arash Torabi and Bentyn, Zofia Antonina and Ignatov, Dmitry and Timofte, Radu},
  booktitle={2025 IEEE/CVF International Conference on Computer Vision Workshops (ICCVW)},
  pages={5723--5733},
  year={2025},
  organization={IEEE}
}

@inproceedings{resnet,
  title={Deep residual learning for image recognition},
  author={He, Kaiming and Zhang, Xiangyu and Ren, Shaoqing and Sun, Jian},
  booktitle={Proceedings of the IEEE conference on computer vision and pattern recognition},
  pages={770--778},
  year={2016}
}

@inproceedings{mobilenetv2,
  title={Mobilenetv2: Inverted residuals and linear bottlenecks},
  author={Sandler, Mark and Howard, Andrew and Zhu, Menglong and Zhmoginov, Andrey and Chen, Liang-Chieh},
  booktitle={Proceedings of the IEEE conference on computer vision and pattern recognition},
  pages={4510--4520},
  year={2018}
}

@article{cifar,
  title={Learning multiple layers of features from tiny images},
  author={Krizhevsky, Alex and Hinton, Geoffrey and others},
  year={2009},
  publisher={Toronto, ON, Canada}
}

@inproceedings{svhn,
  title={Reading digits in natural images with unsupervised feature learning},
  author={Netzer, Yuval and Wang, Tao and Coates, Adam and Bissacco, Alessandro and Wu, Baolin and Ng, Andrew Y and others},
  booktitle={NIPS workshop on deep learning and unsupervised feature learning},
  volume={2011},
  number={2},
  pages={4},
  year={2011},
  organization={Granada}
}

@article{pytorch,
  title={Pytorch: An imperative style, high-performance deep learning library},
  author={Paszke, Adam and Gross, Sam and Massa, Francisco and Lerer, Adam and Bradbury, James and Chanan, Gregory and Killeen, Trevor and Lin, Zeming and Gimelshein, Natalia and Antiga, Luca and others},
  journal={Advances in neural information processing systems},
  volume={32},
  year={2019}
}

@misc{torchvision,
  author = {{TorchVision maintainers and contributors}},
  title  = {{TorchVision}: PyTorch's Computer Vision Library},
  year   = {2016},
  note   = {\url{https://github.com/pytorch/vision}}
}

@misc{olympiccoder,
  author = {{Hugging Face}},
  title  = {{OlympicCoder-7B}},
  year   = {2025},
  note   = {\url{https://huggingface.co/open-r1/OlympicCoder-7B}}
}

@article{qwencoder,
  title={Qwen2.5-coder technical report},
  author={Hui, Binyuan and Yang, Jian and Cui, Zeyu and Yang, Jiaxi and Liu, Dayiheng and Zhang, Lei and Liu, Tianyu and Zhang, Jiajun and Yu, Bowen and Lu, Keming and others},
  journal={arXiv preprint arXiv:2409.12186},
  year={2024}
}

@article{instructgpt,
  title={Training language models to follow instructions with human feedback},
  author={Ouyang, Long and Wu, Jeff and Jiang, Xu and Almeida, Diogo and Wainwright, Carroll L and Mishkin, Pamela and Zhang, Chong and Agarwal, Sandhini and Slama, Katarina and Ray, Alex and others},
  journal={arXiv preprint arXiv:2203.02155},
  year={2022}
}

@inproceedings{flan,
  title={The flan collection: Designing data and methods for effective instruction tuning},
  author={Longpre, Shayne and Hou, Le and Vu, Tu and Webson, Albert and Chung, Hyung Won and Tay, Yi and Zhou, Denny and Le, Quoc V and Zoph, Barret and Wei, Jason and others},
  booktitle={International conference on machine learning},
  pages={22631--22648},
  year={2023},
  organization={PMLR}
}

@article{codex,
  title={Evaluating large language models trained on code},
  author={Chen, Mark and Tworek, Jerry and Jun, Heewoo and Yuan, Qiming and Pinto, Henrique Ponde De Oliveira and Kaplan, Jared and Edwards, Harri and Burda, Yuri and Joseph, Nicholas and Brockman, Greg and others},
  journal={arXiv preprint arXiv:2107.03374},
  year={2021}
}

@article{kojima2022,
  title={Large language models are zero-shot reasoners},
  author={Kojima, Takeshi and Gu, Shixiang Shane and Reid, Machel and Matsuo, Yutaka and Iwasawa, Yusuke},
  journal={Advances in neural information processing systems},
  volume={35},
  pages={22199--22213},
  year={2022}
}

@article{sahoo2025prompt,
  title={A systematic survey of prompt engineering in large language models: Techniques and applications},
  author={Sahoo, Pranab and Singh, Ayush Kumar and Saha, Sriparna and Jain, Vinija and Mondal, Samrat and Chadha, Aman},
  journal={arXiv preprint arXiv:2402.07927},
  year={2024}
}

@inproceedings{successivehalving,
  title={Non-stochastic best arm identification and hyperparameter optimization},
  author={Jamieson, Kevin and Talwalkar, Ameet},
  booktitle={Artificial intelligence and statistics},
  pages={240--248},
  year={2016},
  organization={PMLR}
}

@article{hyperband,
  title={Hyperband: A novel bandit-based approach to hyperparameter optimization},
  author={Li, Lisha and Jamieson, Kevin and DeSalvo, Giulia and Rostamizadeh, Afshin and Talwalkar, Ameet},
  journal={The journal of machine learning research},
  volume={18},
  number={1},
  pages={6765--6816},
  year={2017},
  publisher={JMLR. org}
}

@inproceedings{bohb,
  title={BOHB: Robust and efficient hyperparameter optimization at scale},
  author={Falkner, Stefan and Klein, Aaron and Hutter, Frank},
  booktitle={International conference on machine learning},
  pages={1437--1446},
  year={2018},
  organization={PMLR}
}

@article{egele2024,
  title={The unreasonable effectiveness of early discarding after one epoch in neural network hyperparameter optimization},
  author={Egele, Romain and Mohr, Felix and Viering, Tom and Balaprakash, Prasanna},
  journal={Neurocomputing},
  volume={597},
  pages={127964},
  year={2024},
  publisher={Elsevier}
}

@inproceedings{zerocost,
  author    = {Abdelfattah, Mohamed S. and Mehrotra, Abhinav and Dudziak, {\L}ukasz and Lane, Nicholas D.},
  title     = {Zero-Cost Proxies for Lightweight {NAS}},
  booktitle = ICLR,
  year      = {2021}
}

@inproceedings{mellor2021,
  title={Neural architecture search without training},
  author={Mellor, Joe and Turner, Jack and Storkey, Amos and Crowley, Elliot J},
  booktitle={International conference on machine learning},
  pages={7588--7598},
  year={2021},
  organization={PMLR}
}

@article{ru2021,
  title={Speedy performance estimation for neural architecture search},
  author={Ru, Robin and Lyle, Clare and Schut, Lisa and Fil, Miroslav and van der Wilk, Mark and Gal, Yarin},
  journal={Advances in Neural Information Processing Systems},
  volume={34},
  pages={4079--4092},
  year={2021}
}

@article{nasbench201,
  title={Nas-bench-201: Extending the scope of reproducible neural architecture search},
  author={Dong, Xuanyi and Yang, Yi},
  journal={arXiv preprint arXiv:2001.00326},
  year={2020}
}

@inproceedings{designbench,
  title={Design-bench: Benchmarks for data-driven offline model-based optimization},
  author={Trabucco, Brandon and Geng, Xinyang and Kumar, Aviral and Levine, Sergey},
  booktitle={International Conference on Machine Learning},
  pages={21658--21676},
  year={2022},
  organization={PMLR}
}

@InProceedings{lemur2,
  title={{LEMUR} 2: Unlocking Neural Network Diversity for {AI}},
	 author={Uzun, Tolgay Atinc and Khalid, Waleed and Din, Saif U and Mulukuledu, Sai Revanth and Singh, Akashdeep and Vysyaraju, Chandini and Duvvuri, Raghuvir and Goyal, Avi and Lukhi, Yashkumar Rajeshbhaiand and Hussain, Muhammad Ahsan and Jesani, Krunal and Shrestha, Usha and Mittal, Yash and Kochnev, Roman and Kadam, Pritam and Ikram, Mohsin and Moradiya, Harsh Rameshbhai and Arslanian, Alice and Ignatov, Dmitry and Timofte, Radu},
	booktitle={Proceedings of the IEEE/CVF Conference on Computer Vision and Pattern Recognition Workshops (CVPRW)},
	pages = {3291--3300},		
	year={2026},
	month={June}
}

@InProceedings{vysyaraju2025fewshot,
  	title = {Enhancing {LLM}-Based Neural Network Generation: Few-Shot Prompting and Efficient Validation for Automated Architecture Design},
	author = {Raghuvir Duvvuri and Chandini Vysyaraju and Avi Goyal and Dmitry Ignatov and Radu Timofte},
	booktitle={Proceedings of the IEEE/CVF Conference on Computer Vision and Pattern Recognition Workshops (CVPRW)},	
	pages = {3242--3251},		
	year={2026},
	month={June}
}

@article{aboudeshish2025augmentest,
  title={AUGMENTGEST: Can random data cropping augmentation boost gesture recognition performance?},
  author={Aboudeshish, Nada and Ignatov, Dmitry and Timofte, Radu},
  journal={arXiv preprint arXiv:2506.07216},
  year={2025}
}

@InProceedings{khalid2026creativity,
  title={From Memorization to Creativity: {LLM} as a Designer of Novel Neural Architectures},
	author={Khalid, Waleed and Ignatov, Dmitry and Timofte, Radu},
	booktitle={Proceedings of the IEEE/CVF Conference on Computer Vision and Pattern Recognition Workshops (CVPRW)},	
	pages = {3252--3261},	
	year={2026},
	month={June}
}

@article{bilalli2018,
  title={Intelligent assistance for data pre-processing},
  author={Bilalli, Besim and Abell{\'o}, Alberto and Aluja-Banet, Tom{\`a}s and Wrembel, Robert},
  journal={Computer Standards \& Interfaces},
  volume={57},
  pages={101--109},
  year={2018},
  publisher={Elsevier}
}

@article{schulhoff2025,
  title={The prompt report: A systematic survey of prompt engineering techniques},
  author={Schulhoff, Sander and Ilie, Michael and Balepur, Nishant and Kahadze, Konstantine and Liu, Amanda and Si, Chenglei and Li, Yinheng and Gupta, Aayush and Han, HyoJung and Schulhoff, Sevien and others},
  journal={arXiv preprint arXiv:2406.06608},
  year={2024}
}

@article{acecoder,
  title={Acecoder: Utilizing existing code to enhance code generation},
  author={Li, Jia and Zhao, Yunfei and Li, Yongmin and Li, Ge and Jin, Zhi},
  journal={arXiv preprint arXiv:2303.17780},
  year={2023}
}

@article{lail,
  title={Large language model-aware in-context learning for code generation},
  author={Li, Jia and Tao, Chongyang and Lio, Jia and Li, Ge and Jin, Zhi and Zhang, Huangzhao and Fang, Zheng and Liu, Fang},
  journal={ACM Transactions on Software Engineering and Methodology},
  volume={34},
  number={7},
  pages={1--33},
  year={2025},
  publisher={ACM New York, NY}
}

@misc{parthasarathy2024,
      title={The Ultimate Guide to Fine-Tuning LLMs from Basics to Breakthroughs: An Exhaustive Review of Technologies, Research, Best Practices, Applied Research Challenges and Opportunities}, 
      author={Venkatesh Balavadhani Parthasarathy and Ahtsham Zafar and Aafaq Khan and Arsalan Shahid},
      year={2024},
      eprint={2408.13296},
      archivePrefix={arXiv},
      primaryClass={cs.LG},
      url={https://arxiv.org/abs/2408.13296}, 
}

@article{bai2022rlhf,
  author  = {Bai, Yuntao and Jones, Andy and Ndousse, Kamal and Askell, Amanda and Chen, Anna and DasSarma, Nova and Drain, Dawn and Fort, Stanislav and Ganguli, Deep and Henighan, Tom and Joseph, Nicholas and Kadavath, Saurav and Kernion, Jackson and Conerly, Tom and El-Showk, Sheer and Elhage, Nelson and Hatfield-Dodds, Zac and Hernandez, Danny and Hume, Tristan and Johnston, Scott and Kravec, Shauna and Lovitt, Liane and Nanda, Neel and Olsson, Catherine and Amodei, Dario and Brown, Tom and Clark, Jack and McCandlish, Sam and Olah, Chris and Mann, Ben and Kaplan, Jared},
  title   = {Training a Helpful and Harmless Assistant with Reinforcement Learning from Human Feedback},
  journal = {arXiv preprint arXiv:2204.05862},
  year    = {2022}
}

@article{mumuni2025autodata,
  title={Automated data processing and feature engineering for deep learning and big data applications: a survey},
  author={Mumuni, Alhassan and Mumuni, Fuseini},
  journal={Journal of Information and Intelligence},
  volume={3},
  number={2},
  pages={113--153},
  year={2025},
  publisher={Elsevier}
}

@misc{zhao2021datasetcondensationgradientmatching,
      title={Dataset Condensation with Gradient Matching}, 
      author={Bo Zhao and Konda Reddy Mopuri and Hakan Bilen},
      year={2021},
      eprint={2006.05929},
      archivePrefix={arXiv},
      primaryClass={cs.CV},
      url={https://arxiv.org/abs/2006.05929}, 
}

@InProceedings{ABrain.Analog,
	title={Curating Same-Family Neural Networks for	LLM-Guided Model Improvement: A Controlled Case Study},
	author={Baghel, Kabir Dev Paul and Timofte, Radu and Ignatov, Dmitry},	
	booktitle = {Computer Vision -- ECCV 2026 Workshops},
	series    = {Lecture Notes in Computer Science},
	publisher = {Springer},
	address   = {Cham},
	year	  = {2026},
	note      = {To appear}
}

@article{ABrain.Converg,
	author  = {Santosh Premi Adhikari and Radu Timofte and Dmitry Ignatov},
	title   = {Convergence Theory for Iterative {LLM}-Based Neural Architecture Search: A Parametric Cross-Entropy Framework with Closed-Form Proxy Reliability},
	journal = {arXiv preprint},
	volume  = {arXiv:2605.30103},  
	year    = {2026}
}

@article{ABrain.Delta,
  author  = {Santosh Premi Adhikari and Radu Timofte and Dmitry Ignatov},
  title   = {Delta-based neural architecture search: {LLM} fine-tuning via code diffs},
  journal = {arXiv preprint},
  volume  = {arXiv:2605.04903},  
  year    = {2026}
}

@article{ABrain.Similarity,
	title={Similarity-Guided Curriculum Fine-Tuning of {LLMs} for Neural Architecture Synthesis},
	author={Vijayakumar, Anujaya and Timofte, Radu and Ignatov, Dmitry},	
	journal = {arXiv preprint},
	volume  = {arXiv:2607.11591},
	url = {https://arxiv.org/abs/2607.11591},
	year={2026}
}
}

\end{document}